\pdfoutput=1
\documentclass{article} %
\usepackage{iclr2027_conference,times}

\usepackage{amsmath,amsfonts,bm}

\def\eqref#1{equation~\ref{#1}}

\def\1{\bm{1}}

\DeclareMathAlphabet{\mathsfit}{\encodingdefault}{\sfdefault}{m}{sl}
\SetMathAlphabet{\mathsfit}{bold}{\encodingdefault}{\sfdefault}{bx}{n}

\usepackage{hyperref}       %
\usepackage{url}            %
\usepackage{booktabs}       %
\usepackage{amsfonts}       %
\usepackage{nicefrac}       %
\usepackage{microtype}      %
\usepackage{xcolor}         %
\usepackage{amsmath}
\usepackage{fix-cm}
\usepackage{wrapfig}
\usepackage{graphicx}
\usepackage{cleveref}
\usepackage{multirow}
\usepackage[table]{xcolor}
\usepackage{caption}
\usepackage{algorithm}
\usepackage{algpseudocode}
\usepackage{tabularx}
\usepackage{placeins}
\definecolor{refgray}{RGB}{235,242,250}
\definecolor{closedgray}{gray}{0.88}
\usepackage{xcolor} %
\newcommand{\reftext}[1]{\textcolor[gray]{0.6}{#1}}
\newcommand{\closedtext}[1]{\textcolor[RGB]{150,170,190}{#1}}

\newcommand{\algo}{SpatialCORE}

\title{SpatialCORE: Confidence-Aware \\Grounded Spatial Reasoning \\in Large Vision--Language Models}

\author{Rafi Ibn Sultan$^{1}$ \quad Xiangyu Zhou$^{1}$ \quad Md. Sajid Alam Chowdhury$^{1}$ \quad \\
\textbf{Chengyin Li$^{2}$} \quad \textbf{Prashant Khanduri$^{1}$} \quad \textbf{Marco Brocanelli$^{3}$} \quad \textbf{Dongxiao Zhu}$^{\textbf{1,4}}$\\
$^{1}$\small Department of Computer Science, Wayne State University \quad\\
$^{2}$\small Department of Radiation Oncology, Henry Ford Health \\
$^{3}$\small Department of Electrical and Computer Engineering, The Ohio State University\quad
\\$^{4}$\small Institute for AI and Data Science, Wayne State University\\
}

\iclrfinalcopy %

\begin{document}

\maketitle
\lhead{Preprint}

\begin{abstract}
Large Vision-Language Models (LVLMs) have made remarkable progress across visual
perception tasks, yet spatial reasoning remains a persistent weakness, especially
for questions that require reasoning over visual space. Recent spatial-reasoning
methods incorporate generated grounding, where models predict bounding boxes,
masks, or other localization outputs for task-relevant objects as part of their
reasoning trace. However, these approaches typically optimize final-answer correctness alone,
allowing correct answers to be rewarded even when the model does not reason
from confidently localized task-relevant objects. We introduce \textbf{SpatialCORE}
\textbf{(Spatial}ly \textbf{CO}nfident \textbf{RE}asoning), a post-training framework that turns the model's own confidence in generated grounding into a learning signal for spatial reasoning. Its central idea is to reinforce grounding that is both accurate and confident, encouraging the model to reason from confidently localized task-relevant objects. SpatialCORE realizes this through a \textit{self-regulating spatial reward} that weights each predicted grounding, i.e., bounding box's matching quality by its coordinate-token confidence. An answer gate further ties grounding optimization to final-answer correctness. SpatialCORE achieves state-of-the-art results among
open-source and specialized spatial reasoning models across diverse benchmarks,
and transfers effectively in zero-shot settings to unseen data distributions. The
source code is available at \url{https://github.com/rafiibnsultan/SpatialCORE}.
\end{abstract}

\section{Introduction}
\label{sec:intro}

Large Vision--Language Models (LVLMs)~\cite{alayrac2022flamingo,li2023blip,liu2023visual,
dai2023instructblip,bai2023qwen} have achieved strong performance on multimodal perception tasks, yet they remain unreliable when reasoning about spatial structure~\cite{qi2025beyond,ranasinghe2024learning,
zhou2025robotracer,xu2026thinking,yang2025thinking,song2025robospatial}. Even with accurate object identification, they often struggle to understand spatial arrangements and relationships~\cite{liu2026spatial,zhang2025mllms,yu2025far,sun2025spacevista,liu2025can}. Spatial
reasoning requires the model to move beyond recognizing scene elements and
construct a coherent understanding of the environment through fine-grained
relationships among objects~\cite{yang2025visual,gholami2025spatial,batra2025spatialthinker,lee2025perspective}. This gap between object recognition and spatial understanding limits the use
of LVLMs in real-world settings such as robotics~\cite{goral2024seeing},
autonomous driving~\cite{wu2024vsp}, and pedestrian
assistance~\cite{sultan2026walkgpt}, where correct decisions depend on
reasoning about spatial relationships rather than recognizing objects alone.

Efforts to address this weakness have taken different forms. Some methods rely on
external guidance, such as user-provided points, regions, or spatial anchors, to
indicate where task-relevant objects are located or how they should be compared
spatially~\cite{Pothiraj_2025_ICCV,cheng2024spatialrgpt,cai2025depthlm,
gholami2025spatial,shen2025fine}. While effective, these methods depend on such
guidance and do not directly train the model to reason spatially on its own.
A second line post-trains LVLMs on spatial reasoning tasks by supervising
grounding predictions, such as segmentation masks or BBoxes, as model
outputs~\cite{ning2025enhancing,chen2024spatialvlm,yang2025visual,
ranasinghe2024learning}. While this improves localization as a standalone
training objective, the grounding remains separate from the reasoning trace. Most recently, inspired by ``Thinking with Images''~\cite{openai2025thinking},
models have been encouraged to incorporate generated grounding during
reasoning~\cite{wu2025reinforcing,zheng2025deepeyes,ma2026thinking,
batra2025spatialthinker,chen2025sifthinker,li2025spatialladder}. In these
methods, generated grounding indicates the task-relevant objects the model uses while
reasoning, typically through bounding boxes (BBoxes), masks, or similar localization cues.
However, neither answer correctness nor localization quality alone explicitly captures the model's confidence in generated grounding.

\begin{wrapfigure}{r}{0.55\textwidth}
    \vspace{-5mm}
    \includegraphics[width=\linewidth]{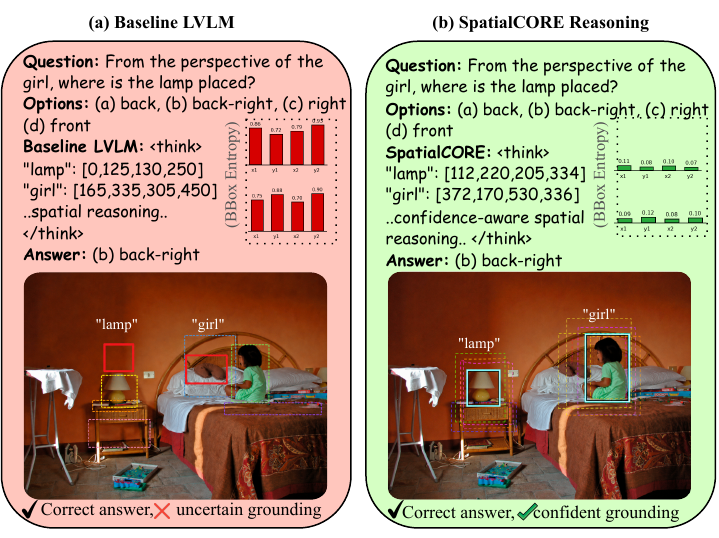}
    \vspace{-5mm}
    \caption{\small Correct final answers do not necessarily imply confident grounding. Although both models answer correctly, \textbf{(a)} the \textbf{baseline LVLM}
    produces \textcolor{red}{high-entropy} predicted BBoxes, with dashed
    candidate BBoxes spread across off-target locations. \textbf{(b)} Our \textbf{\algo}
    produces \textcolor{green}{lower-entropy} predicted BBoxes, with dashed
    candidate BBoxes concentrated around the selected BBoxes, and more confidently
    localizing the task-relevant objects. Solid BBoxes denote the predicted
    BBoxes in the reasoning trace; dashed BBoxes denote candidate BBoxes reflected by
    spatial uncertainty.}
    \label{fig:figure1}
    \vspace{-3mm}
\end{wrapfigure}

This is especially problematic for spatial reasoning, where generated grounding
should localize the task-relevant objects needed to compare positions,
distances, and relationships. Most training objectives primarily reward
correct final answers~\cite{li2025spatialladder,ma2026thinking,
batra2025spatialthinker}, even when reasoning includes predicted BBoxes
for these objects. Some of these methods add spatial or trajectory-level rewards,
but still do not account for how confidently the model generates its
grounding. A predicted BBox's coordinate-token uncertainty provides a
way to estimate this confidence during reasoning. \Cref{fig:figure1}a illustrates this distinction: the baseline correctly answers
``back-right'' when asked where the lamp is relative to the girl, yet
generates high-uncertainty BBoxes that poorly localize both objects.
Because the answer is correct, a final-answer reward reinforces the
entire trajectory, including its uncertain grounding. This motivates
our central question: \emph{Can spatial reasoning in LVLMs be improved
by learning from the confidence of their own generated grounding?}

To address this, we propose \textbf{Spatial}ly \textbf{CO}nfident
\textbf{RE}asoning~\textbf{(\algo)}, a post-training framework that enhances spatial reasoning in LVLMs by learning to ground with confidence. As illustrated in \Cref{fig:figure1}b, \algo\ encourages correct answers to be supported by low-uncertainty BBoxes for task-relevant objects. Its \textit{self-regulating spatial reward} uses the model's own confidence, estimated from BBox coordinate uncertainty, to weight geometric overlap with reference BBoxes. This makes grounding confidence a learning signal even among trajectories reaching the same correct answer. An answer gate further couples spatial and answer rewards by scaling the spatial reward according to final-answer correctness.

Our contributions are summarized as follows:
\begin{itemize}
    \item We introduce \algo, a novel post-training framework that makes the model's own confidence in generated grounding an explicit learning signal for spatial reasoning in LVLMs.

    \item We propose a self-regulating spatial reward that evaluates generated grounding through both localization quality and coordinate-token confidence, with an answer gate connecting grounding optimization to final-answer correctness.

    \item We demonstrate state-of-the-art results among open-source and specialized spatial reasoning models, with effective zero-shot generalization. Controlled ablations establish the value of confidence weighting, while grounding analyses reveal improved alignment between confidence and localization quality.
\end{itemize}

\section{Related Works}
\label{sec:related_works}

\noindent\textbf{Externally Guided Spatial Grounding.}
A common strategy for improving spatial reasoning in LVLMs is to provide
explicit spatial anchors as input. Points, regions, masks, or referenced objects
are supplied with the query to direct the model toward task-relevant
objects~\cite{bigverdi2025perception,yang2025visual,cheng2024spatialrgpt,
shen2025fine,ma2025spatialllm, cai2025depthlm}. These approaches are effective when reliable anchors are
available, but their dependence on inference-time guidance limits open-ended use:
when anchors are absent or ambiguous, the model must still identify relevant
objects on its own. Thus, spatial reasoning may not transfer to settings without
external anchors.

\noindent\textbf{Geometry-Enhanced Visual Understanding.}
Another line improves spatial reasoning by adding geometric cues to the visual
representation. These methods use depth maps, point clouds, segmentation masks,
or multi-view geometry to encode scene layout and spatial
relationships~\cite{liu2025ssr,chen2024ll3da,hu2025g,wang2025n3d,
wan2025eaglevision,cai2025spatialbot,chensd,sultan2026walkgpt,ning2025enhancing,
daxberger2503mm,hong20233d,wu2025spatial,xu2026s,zhao2025spacemind,chen2026think,zhou2026learning}. Such representations can improve spatial perception, but they mainly change what
the model observes, not how it learns to generate and use grounding during
reasoning. These representations do not by themselves specify how grounding confidence should affect the post-training reward.

\noindent\textbf{Inference-Time Reasoning Scaffolds.}
Spatial reasoning can also be improved at inference time by structuring the
model's response without post-training. These methods guide reasoning through
cognitive maps, scene graphs, perspective-aware representations, compositional
prompting, or related scaffolds~\cite{liao2024reasoning,gholami2025spatial,
lee2025perspective,ma2024spatialpin,mitra2024compositional,yang2025thinking},
and may also intervene at decoding time~\cite{chen2025spatial,
huang2025spatial,yangrasp}. While training-free, their gains are often tied to specific
task formats, prompts, scaffolds, or decoding procedures, leaving the model's
spatial reasoning behavior unoptimized.

\noindent\textbf{Training-Based Spatial Reasoning.}
A more direct strategy is to optimize spatial reasoning during training \cite{kancheti2026faithful,chen2026spacetools,li2026star}. One line
enriches the reasoning trace with spatial content, such as generated grounding or
other task-relevant spatial outputs~\cite{ma2026thinking,chen2025sifthinker,
batra2025spatialthinker,wang2025svqa}. Inspired by ``Thinking with
Images''~\cite{openai2025thinking}, a related line further supervises generated
grounding in the reasoning trace with RL-based objectives that reward correct localization of
task-relevant objects~\cite{wu2025reinforcing,zheng2025deepeyes,xu2025visual,
sarch2025grounded,zhao2025embodied,li2025spatialladder}. These approaches emphasize what grounding is produced and whether it is correct. \algo\ introduces the model's confidence in producing that grounding as an additional learning signal, training spatial reasoning through both grounding quality and certainty.

\section{Method}
\label{sec:method}
\begin{figure*}[t]
\centering
\includegraphics[width=1.0\textwidth]{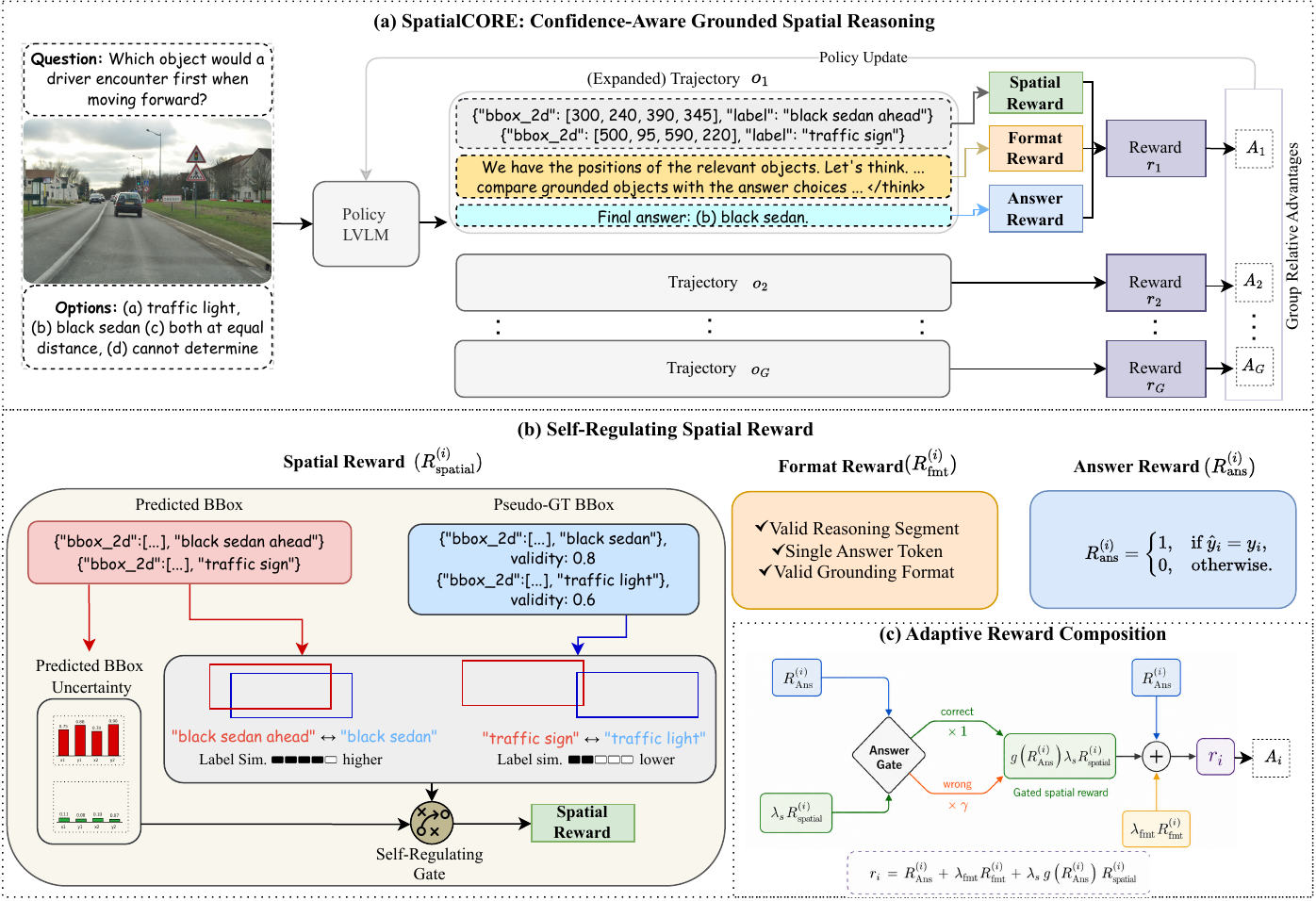}
\caption{\small Overview of \textbf{\algo}. \textbf{(a)} The LVLM policy samples
trajectories comprising a reasoning trace with generated grounding expressed as
bounding boxes (BBoxes), followed by a final answer. \textbf{(b)} The \textit{self-regulating spatial reward} uses predicted BBox
coordinate-token uncertainty to estimate the grounding confidence, which
then weights each BBox's matching quality. Predicted BBoxes are matched to pseudo-GT
BBoxes using geometric overlap, label similarity, and pseudo-GT validity. For
example, \textit{black sedan ahead} and \textit{black sedan} have high label
similarity, while higher pseudo-GT validity, such as \textit{black sedan,
validity: 0.8}, gives the match more weight. \textbf{(c)} Spatial, format, and
answer rewards are composed through an \textit{answer gate} to produce
trajectory-level rewards, which are used to compute group-relative advantages for
policy update.}

\label{fig:figure2}
\vspace{-10pt}
\end{figure*}

We develop \algo~(\Cref{fig:figure2}a), a post-training framework that improves
spatial reasoning in LVLMs through confidence-aware grounding. Built on Group
Relative Policy Optimization (GRPO)~\cite{shao2024deepseekmath}, \algo\
optimizes sampled reasoning trajectories using a \textit{self-regulating spatial
reward} that weights generated grounding by model confidence, encouraging
reasoning from confidently localized task-relevant objects. The reward measures
uncertainty in predicted bounding-box coordinates (\Cref{fig:figure2}b) and is
combined with format and answer rewards through an \textit{answer gate}. The
resulting trajectory-level rewards are normalized within each rollout group for
the GRPO policy update (\Cref{fig:figure2}c).

\subsection{Problem Formulation}

Given an image $I$ and a spatial reasoning query $q$, a LVLM policy
$\pi_\theta$ generates a trajectory $o=(x_1,\ldots,x_T)$ consisting of a
reasoning trace with generated grounding tokens that specify predicted bounding boxes (BBoxes), followed by a final answer. Based on the next-token distribution $\pi_\theta(\cdot \mid x_{<t}, I, q)$, the trajectory likelihood $\pi_\theta(o \mid I,q)$ is defined as
\begin{equation}
\pi_\theta(o \mid I,q) = \prod_{t=1}^{T} \pi_\theta(x_t \mid x_{<t}, I, q).
\end{equation}

For each input $(I,q)$, the policy performs a GRPO rollout by sampling a group of $G$ trajectories $\{o_i\}_{i=1}^{G}$, as illustrated in \Cref{fig:figure2}a. The rollout assigns each trajectory a reward $r_i$ and computes the corresponding group-relative advantage $A_i$, which then enters the GRPO policy objective:
\begin{equation}
r_i = R(o_i), \qquad
A_i = \frac{r_i - \frac{1}{G}\sum_{j=1}^{G} r_j}{\sigma_G + \varepsilon},
\end{equation}
\noindent where $\sigma_G$ is the standard deviation of the group rewards and $\varepsilon$ is a small constant for numerical stability.

\subsection{Self-Regulating Spatial Reward}
The \textit{self-regulating spatial reward} operates on predicted BBoxes
within each reasoning trajectory. It estimates BBox confidence from the current policy's coordinate-token uncertainty, so low-uncertainty BBoxes contribute more to the
reward, while high-uncertainty BBoxes contribute less
(\Cref{fig:figure2}b). To obtain pseudo-ground-truth bounding boxes (pseudo-GT BBoxes), we use a
Referring Expression Comprehension (REC) model (e.g. Grounding DINO ~\cite{liu2024grounding}) to
localize task-relevant objects. During reward computation, each pseudo-GT BBox is weighted by its REC validity, so a higher-validity localization such as \textit{black sedan} with validity \(0.8\) contributes more than a lower-validity localization such as \textit{traffic light} with validity \(0.6\).

\subsubsection{Confidence-Aware Spatial Reward}

\noindent\textbf{Pseudo-GT-Guided Spatial Matching.}
During each trajectory $o_i$, the
LVLM policy generates predicted BBoxes for task-relevant objects referenced in the question and answer options as part of the reasoning trace. These
predicted BBoxes are evaluated against precomputed pseudo-GT BBoxes from a
REC model $\mathcal{G}$. For each image-question pair, $\mathcal{G}$
localizes the extracted objects into tuples
$(b_k^{\mathrm{gt}}, \ell_k^{\mathrm{gt}}, v_k)$, where
$b_k^{\mathrm{gt}}$ is the pseudo-GT BBox, $\ell_k^{\mathrm{gt}}$ is the
corresponding label of a task-relevant object, and $v_k \in [0,1]$ is the validity produced by $\mathcal{G}$ for the localized pseudo-GT BBox.

As illustrated in \Cref{fig:figure2}b, each trajectory may generate multiple
predicted BBoxes, so matching them to pseudo-GT BBoxes cannot rely on geometric
overlap alone. A generated label such as \textit{black sedan ahead} should match
\textit{black sedan} more strongly than a mismatched label such as
\textit{traffic sign}. We therefore compute a pairwise BBox reward for each predicted BBox $b_j$ with object label $\ell_j$ against each pseudo-GT BBox $b_k^{\mathrm{gt}}$, using
both geometric overlap and label similarity:
\begin{equation}
R_{\text{BBox}}^{(j,k)} =
\left(
w_{\text{iou}} \cdot \max\left(0, \mathrm{IoU}(b_j, b_k^{\mathrm{gt}}) - \tau_{\text{iou}}\right)
+ w_{\text{label}} \cdot \mathrm{Sim}(\ell_j, \ell_k^{\mathrm{gt}})
\right) \cdot v_k,
\end{equation}
where $w_{\mathrm{iou}} + w_{\mathrm{label}} = 1$ and $\tau_{\mathrm{iou}}$
is an IoU margin. The clipped IoU term suppresses weak geometric overlap, while
$\mathrm{Sim}(\ell_j,\ell_k^{\mathrm{gt}})$ measures label similarity using
cosine similarity between semantic label representations. The pseudo-GT
validity $v_k$ scales the pairwise BBox reward, assigning a larger weight to
higher-validity pseudo-GT BBoxes and reducing the influence of noisier ones.
This makes the spatial reward depend more on reliable pseudo-GT BBoxes during
matching. We then use Hungarian matching~\cite{kuhn1955hungarian} to obtain the
optimal one-to-one assignment between predicted BBoxes and pseudo-GT BBoxes:
\begin{equation}
\mathcal{M}_i^*
=
\arg\max_{\mathcal{M}_i}
\sum_{(j,k)\in\mathcal{M}_i} R_{\text{BBox}}^{(j,k)}.
\end{equation}
The matched pairs in $\mathcal{M}_i^*$ are then used to compute the confidence-weighted spatial reward.

\noindent\textbf{BBox Coordinate-Token Uncertainty.}
We estimate the confidence of generated grounding from the tokens that
produce each predicted BBox. A BBox is emitted as
\texttt{"bbox\_2d": [x\_1, y\_1, x\_2, y\_2]}, with each coordinate
generated autoregressively as digit tokens that determine its location;
their uncertainty therefore estimates confidence in the predicted BBox.

Let \(\mathcal{C} = \{x_1, y_1, x_2, y_2\}\) denote the BBox coordinates
and \(\mathcal{S}_{i,j,c}\) the digit-token positions of coordinate
\(c \in \mathcal{C}\) of predicted BBox \(b_j\) in trajectory \(o_i\),
excluding brackets, commas, and spaces. Let \(h_{i,t}\) denote the
Shannon entropy of \(P_\theta(\cdot \mid x_{i,<t}, I, q)\) over the full
vocabulary \(\mathcal{V}\), and \(\mathcal{D} \subset \mathcal{V}\) the
set of digit tokens. We define the uncertainty of \(b_j\) as the
normalized entropy averaged over digits within each coordinate, then
over coordinates:
\begin{equation}
H_{i,j} =
\operatorname{clip}_{[0,1]}
\left(
\frac{1}{|\mathcal{C}| \log |\mathcal{D}|}
\sum_{c \in \mathcal{C}}
\frac{1}{|\mathcal{S}_{i,j,c}|}
\sum_{t \in \mathcal{S}_{i,j,c}}
h_{i,t}
\right).
\end{equation}
Full-vocabulary entropy also rises when probability shifts to non-digit
tokens, while \(\log |\mathcal{D}|\), the entropy of a uniform choice
over digits, sets its scale.  Lower \(H_{i,j}\)
indicates higher confidence in \(b_j\).

\noindent\textbf{Confidence-Weighted Spatial Reward.}
We weight each matched predicted BBox by its confidence, estimated from
coordinate-token uncertainty \(H_{i,j}\):
\begin{equation}
C_{i,j} = 1 - H_{i,j}, \qquad
\omega_{i,j} = \beta + (1-\beta)\,C_{i,j},
\end{equation}
where $C_{i,j}$ denotes the confidence of predicted BBox $b_j$, $\beta \in (0,1)$
is a confidence floor, and $(j,k)\in\mathcal{M}_i^*$ denotes a matched
prediction--pseudo-GT pair. The confidence weight gives high-certainty groundings greater
contribution while remaining positive even under high uncertainty. The resulting weighted matching scores collectively define a confidence-weighted measure of matching quality,
\begin{equation}
P_i =
\frac{1}{|\mathcal{M}_i^*|}
\sum_{(j,k)\in\mathcal{M}_i^*}
\omega_{i,j}\,R_{\mathrm{BBox}}^{(j,k)},
\end{equation}
where \(P_i=0\) when no match is found. This term evaluates matched
prediction--pseudo-GT pairs under the one-to-one assignment, so ambiguous or
duplicated assignments cannot inflate the matching quality.

To encourage broader coverage of the pseudo-GT BBoxes for the input \((I,q)\),
denoted by \(\mathcal{P}\), we define a soft recall term weighted by pseudo-GT
validity:
\begin{equation}
\mathrm{Recall}_i
=
\left(\sum_{k \in \mathcal{P}} v_k\right)^{-1}
\sum_{k \in \mathcal{P}} \max_j R_{\mathrm{BBox}}^{(j,k)},
\qquad
R_{\mathrm{spatial}}^{(i)}
=
F_\alpha(P_i,\mathrm{Recall}_i)
=
\frac{(1+\alpha^2)P_i\,\mathrm{Recall}_i}
{\alpha^2 P_i + \mathrm{Recall}_i}.
\end{equation}
If \(\sum_{k \in \mathcal{P}} v_k=0\) or the \(F_\alpha\) denominator is zero,
we set \(R_{\mathrm{spatial}}^{(i)}=0\). Here, \(\mathrm{Recall}_i\) measures
how well the pseudo-GT BBoxes are covered by the set of predictions, rewarding
each task-relevant object that is localized by at least one predicted BBox. The
spatial reward combines matching quality and pseudo-GT coverage through an
\(\alpha\)-weighted harmonic mean, with \(\alpha>1\) mildly emphasizing coverage
and discouraging trajectories that localize only a subset of task-relevant
objects.

\subsubsection{Format and Answer Rewards}

\noindent\textbf{Format Reward.}
We use a format reward to enforce the trajectory structure required for grounded
reasoning. As shown in \Cref{fig:figure2}b, a trajectory is rewarded for three
format properties: a valid reasoning segment, a single final answer token, and a
valid grounding format for predicted BBoxes. Malformed outputs, repeated object
labels, or BBoxes placed outside the reasoning segment reduce the format reward.

\noindent\textbf{Answer Reward.}
The answer reward assigns a unit reward to a correct final answer and zero otherwise:
\begin{equation}
R_{\text{ans}}^{(i)} =
\begin{cases}
1, & \text{if } \hat{y}_i = y_i, \\
0, & \text{otherwise.}
\end{cases}
\end{equation}

\subsection{Adaptive Reward Composition}
We integrate the answer, format, and confidence-aware spatial rewards to assign
a single reward to each sampled trajectory. The spatial term is adaptively modulated by an \textit{answer gate}, so the
spatial reward remains tied to final-answer correctness
(\Cref{fig:figure2}c):
\begin{equation}
r_i
=
R_{\text{ans}}^{(i)}
+
\lambda_{\text{fmt}} R_{\text{fmt}}^{(i)}
+
\lambda_s\, g\!\left(R_{\text{ans}}^{(i)}\right) R_{\text{spatial}}^{(i)},
 \label{eq:reward_compose}
\end{equation}
where \(g(1)=1\) and \(g(0)=\gamma\), with \(\gamma \in (0,1)\). When the final
answer is correct, the full spatial reward is used; when the final answer is
incorrect, positive spatial rewards are reduced by \(\gamma\). This allows
incorrect-answer trajectories to retain partial credit for useful generated grounding.

\begin{table*}[t]
\centering
\small
\caption{\small OmniSpatial \cite{jia2025omnispatial} results across 10 spatial reasoning task categories.
\algo\ is compared against proprietary models, general open-source LVLMs, and
specialized spatial reasoning models. Proprietary models are included as reference
points; the best open-source or specialized result is shown in \textbf{bold}.
Average accuracy is weighted by category sample size.}
\resizebox{\textwidth}{!}{
\begin{tabular}{lccccccccccc}
\toprule
\midrule
\multirow{2}{*}{Method} & \multirow{2}{*}{Average}
& \multicolumn{2}{c}{Dynamic Reasoning}
& \multicolumn{3}{c}{Spatial Interaction}
& \multicolumn{2}{c}{Complex Logic}
& \multicolumn{3}{c}{Perspective Taking} \\
\cmidrule(lr){3-4} \cmidrule(lr){5-7} \cmidrule(lr){8-9} \cmidrule(lr){10-12}
& & \shortstack[c]{Mani-\\pulation} & \shortstack[c]{Motion\\Anal.} & \shortstack[c]{Traffic\\Anal.}
& \shortstack[c]{Loca-\\lization} & \shortstack[c]{Geospa.\\Strategy} & \shortstack[c]{Pattern\\Rec.}
& \shortstack[c]{Geometric\\Reasoning}
& \shortstack[c]{Ego\\Centric} & \shortstack[c]{Allo\\Centric} & \shortstack[c]{Hypo-\\thetical} \\
\hline

\multicolumn{12}{l}{\textbf{Reference Baselines}} \\
\reftext{Random Choice} & \reftext{24.98} & \reftext{24.86} & \reftext{26.30} & \reftext{25.88} & \reftext{23.43} & \reftext{27.27} & \reftext{21.44} & \reftext{24.77} & \reftext{22.55} & \reftext{24.84} & \reftext{25.78} \\
\reftext{Human Evaluation} & \reftext{92.63} & \reftext{94.62} & \reftext{96.07} & \reftext{91.38} & \reftext{95.11} & \reftext{92.15} & \reftext{89.02} & \reftext{85.90} & \reftext{98.53} & \reftext{94.30} & \reftext{90.26} \\
\hline
\multicolumn{12}{l}{\textbf{Proprietary Models}} \\
\closedtext{GPT-4.1-mini} \closedtext{\cite{open2025introducing}} & \closedtext{48.87} & \closedtext{64.32} & \closedtext{56.53} & \closedtext{59.06} & \closedtext{60.19} & \closedtext{56.36} & \closedtext{29.28} & \closedtext{30.19} & \closedtext{72.55} & \closedtext{39.57} & \closedtext{39.28} \\
\closedtext{Gemini-2.5-flash-preview} \closedtext{\cite{team2023gemini}} & \closedtext{52.12} & \closedtext{67.57} & \closedtext{62.72} & \closedtext{68.24} & \closedtext{73.33} & \closedtext{60.91} & \closedtext{38.14} & \closedtext{34.19} & \closedtext{75.49} & \closedtext{35.90} & \closedtext{33.73} \\
\closedtext{o4-mini} \closedtext{\cite{openai2025o3o4mini_system_card}} & \closedtext{52.77} & \closedtext{72.97} & \closedtext{59.83} & \closedtext{60.00} & \closedtext{73.33} & \closedtext{61.82} & \closedtext{34.02} & \closedtext{36.77} & \closedtext{73.53} & \closedtext{40.69} & \closedtext{40.96} \\
\closedtext{Gemini-2.5-flash} \closedtext{\cite{team2023gemini}} & \closedtext{53.16} & \closedtext{70.27} & \closedtext{64.74} & \closedtext{61.18} & \closedtext{72.38} & \closedtext{58.18} & \closedtext{35.05} & \closedtext{36.13} & \closedtext{74.12} & \closedtext{40.96} & \closedtext{32.53} \\
\hline

\multicolumn{12}{l}{\textbf{Open-source Models}} \\

LLaVA-1.5-7B \cite{liu2024improved}& 34.97 & 54.46 & 31.23 & 35.29 & 36.19 & 33.94 & 29.01 & 24.18 & 55.60 & 34.66 & 36.14 \\
LLaVA-OV-7B \cite{li2024llava}& 35.68 & 43.24 & 38.15 & 32.94 & 29.52 & 41.82 & 28.87 & 22.58 & 47.06 & 36.17 & 37.35 \\
InternVL3-8B \cite{zhu2025internvl3}& 41.60 & 52.43 & 40.87 & 48.94 & 51.05 & 44.77 & 24.95 & 28.63 & 64.20 & 38.62 & 40.96 \\
InternVL3-14B \cite{zhu2025internvl3}& 45.94 & 54.32 & \textbf{60.17} & 50.35 & 51.81 & 51.45 & 28.04 & 28.26 & 68.04 & 35.37 & 34.46 \\

Qwen2.5-VL-7B \cite{wang2024qwen2}& 39.18 & 58.38 & 35.09 & 50.12 & 45.33 & 44.00 & 31.13 & 29.42 & 64.51 & 33.19 & 37.35 \\
Gemma-3-12B \cite{gemmateam2025gemma3technicalreport}& 43.71 & 54.05 & 54.91 & 54.12 & 47.62 & 45.45 & 16.49 & 30.32 & 63.73 & 36.70 & 33.73 \\

\hline

\multicolumn{12}{l}{\textbf{Spatial Reasoning Models}} \\

SpaceMantis-13B \cite{chen2024spatialvlm}& 36.36 & 47.03 & 36.59 & 40.94 & 34.86 & 33.09 & 22.27 & 24.39 & 49.22 & 38.25 & 39.28 \\
SpaceQwen2.5-VL-3B \cite{chen2024spatialvlm}& 40.25 & 58.11 & 39.88 & 41.18 & 40.95 & 40.91 & 29.90 & 25.81 & 63.73 & \textbf{38.83} & 39.76 \\
SpaceThinkerQwen2.5VL-3B \cite{chen2024spatialvlm}& 40.42 & 47.84 & 53.06 & 43.29 & 35.43 & 38.73 & 24.33 & 28.00 & 58.04 & 35.11 & 31.08 \\
RoboPoint-vicuna-v1.5-7B-lora \cite{cai2025spatialbot}& 35.85 & 57.03 & 28.61 & 34.82 & 37.33 & 40.55 & 29.90 & 22.71 & 50.20 & 38.72 & 40.96 \\
RoboPoint-vicuna-v1.5-13B \cite{liu2023syncdreamer}& 34.60 & 55.68 & 28.15 & 42.82 & 32.19 & 32.55 & 24.12 & 27.74 & 49.02 & 37.66 & 33.49 \\
VST-RL-7B \cite{yang2025visual}& 41.09 & 56.75 & 43.39 & 44.75 & 46.66 & 42.72 & 25.51 & 28.38 & \textbf{72.54} & 32.89 & 43.37 \\
SoFar-Qwen2.5VL-3B \cite{qi2025sofar}& 45.14 & 56.49 & 51.16 & 54.12 & 53.14 & 52.73 & \textbf{31.75} & 22.88 & 71.60 & 36.56 & 41.69 \\
SpatialLadder-3B \cite{li2025spatialladder}& 40.50 & 59.45 & 39.01 & 50.58 & 48.57 & 42.72 & 26.80 & 23.87 & 71.56 & 35.10 & 39.75 \\

\hline

\multicolumn{12}{l}{\textbf{Backbone and Our Variants}} \\

\algo-4B & 44.68 & 64.68 & 51.73 & 50.58 & \textbf{63.80} & 46.36 & 19.58 & 27.74 & 66.66 & 34.84 & \textbf{44.57} \\
Qwen3-VL-8B-Thinking (Base) \cite{bai2025qwen3} & 43.90 & 57.14 & 54.28 & 36.04 & 54.95 & 48.67 & 24.77 & 23.12 & 71.56 & 31.11 & 38.82 \\
Qwen3-VL-8B-Thinking + GRPO & 45.92 & 61.32 & 54.30 & 55.88 & 58.33 & 46.18 & 25.90 & 25.00 & 72.92 & 36.57 & 42.53 \\
\algo-8B  & \textbf{48.46} & \textbf{64.86} & 56.64 & \textbf{57.64} & 62.85 & \textbf{53.63} & 30.92 & \textbf{36.77} & \textbf{72.54} & 31.91 & 40.96 \\

\midrule
\bottomrule

\end{tabular}
}
\vspace{-5mm}
\label{tab:table1}
\end{table*}

\subsection{Policy Optimization}

The policy is optimized using a GRPO-style clipped objective with KL
regularization against a reference policy $\pi_{\mathrm{ref}}$. Let
$\rho_i(\theta)=
\pi_\theta(o_i \mid I,q)/
\pi_{\theta_{\mathrm{old}}}(o_i \mid I,q)$ denote the importance ratio for
trajectory $o_i$, where $\pi_{\theta_{\mathrm{old}}}$ is the rollout policy used
to sample the current group. Using the group-relative advantages
$\{A_i\}_{i=1}^{G}$ defined above, we optimize the following objective, where
$\delta$ is the clipping threshold and $\eta$ is the KL penalty coefficient:
\begin{equation}
\mathcal{J}(\theta)
=
\mathbb{E}_{(I,q) \sim \mathcal{D},\,\{o_i\}\sim \pi_{\theta_{\mathrm{old}}}}
\!\left[
\frac{1}{G}
\sum_{i=1}^{G}
\min\!\left(
\rho_i(\theta) A_i,
\mathrm{clip}(\rho_i(\theta), 1-\delta, 1+\delta) A_i
\right)
-
\eta\,D_{\mathrm{KL}}\!\left(\pi_\theta \,\|\, \pi_{\text{ref}}\right)
\right].
\end{equation}

\section{Experiments}
\label{sec:experiments}

\begin{table*}[t]
\centering
\small
\caption{\small SpatiaLab~\cite{wasi2026spatialab} results across 6 spatial
reasoning task categories in the zero-shot setting. \algo\ is compared against
proprietary models, general open-source LVLMs, and specialized spatial reasoning
models. Proprietary models are included as reference points; the best open-source
or specialized result is shown in \textbf{bold}. Average accuracy is weighted by
category sample size.}
\resizebox{\textwidth}{!}{
\begin{tabular}{lccccccc}
\toprule
\midrule
\multirow{2}{*}{Method} & \multirow{2}{*}{Average}
& \multicolumn{6}{c}{Question Categories} \\
\cmidrule(lr){3-8}
& & 3D Geom. & Dep. \& Occu. & Orientation & Relat. Posit. & Size \& Scale & Spati. Navig. \\
\hline

\multicolumn{8}{l}{\textbf{Reference Baselines}} \\
\reftext{Random Choice} & \reftext{25.00} & \reftext{25.00} & \reftext{25.00} & \reftext{25.00} & \reftext{25.00} & \reftext{25.00} & \reftext{25.00} \\
\reftext{Human Baseline} & \reftext{87.57} & \reftext{93.70} & \reftext{74.13} & \reftext{91.58} & \reftext{91.51} & \reftext{88.89} & \reftext{87.76} \\
\hline
\multicolumn{8}{l}{\textbf{Proprietary Models}} \\
\closedtext{GPT-4o-mini} \closedtext{\cite{hurst2024gpt}} & \closedtext{46.50} & \closedtext{47.06} & \closedtext{39.00} & \closedtext{47.03} & \closedtext{47.17} & \closedtext{49.60} & \closedtext{49.79} \\
\closedtext{Gemini-2.5-Flash} \closedtext{\cite{team2023gemini}} & \closedtext{48.29} & \closedtext{44.96} & \closedtext{48.26} & \closedtext{48.02} & \closedtext{56.13} & \closedtext{42.46} & \closedtext{51.05} \\
\closedtext{Claude 3.5 Haiku} \closedtext{\cite{anthropic2024model}} & \closedtext{42.93} & \closedtext{42.44} & \closedtext{42.08} & \closedtext{46.53} & \closedtext{46.23} & \closedtext{35.71} & \closedtext{45.99} \\
\closedtext{Mistral Medium 3.1} \closedtext{\cite{mistralai2025medium31}} & \closedtext{47.93} & \closedtext{46.64} & \closedtext{49.81} & \closedtext{47.52} & \closedtext{61.79} & \closedtext{41.67} & \closedtext{41.77} \\
\closedtext{Kimi-VL-A3B-Thinking-2506} \closedtext{\cite{team2025kimi}} & \closedtext{42.71} & \closedtext{42.86} & \closedtext{41.31} & \closedtext{40.59} & \closedtext{51.42} & \closedtext{39.68} & \closedtext{41.35} \\

\hline

\multicolumn{8}{l}{\textbf{Open-source Models}} \\

InternVL3.5-1B \cite{wang2025internvl3}& 31.64 & 33.61 & 32.43 & 23.27 & 37.26 & 31.75 & 30.80 \\
InternVL3.5-2B \cite{wang2025internvl3}& 33.71 & 34.03 & 31.66 & 31.68 & 40.57 & 32.54 & 32.49 \\
Qwen2.5-VL-3B-Instruct \cite{wang2024qwen2}& 41.43 & 41.18 & 35.52 & 46.04 & 40.09 & \textbf{47.22} & 39.24 \\

InternVL3.5-4B \cite{wang2025internvl3}& 43.29 & 42.86 & 42.86 & 42.08 & 54.72 & 36.51 & 42.19 \\
Gemma-3-4B-it \cite{gemmateam2025gemma3technicalreport}& 40.57 & 43.70 & 34.36 & 46.53 & 45.75 & 37.30 & 37.97 \\
LLaVA-1.5-7B \cite{liu2024improved} & 38.64 & 40.33 & 34.74 & 31.68 & 40.56 & 40.07 & 43.88 \\
Qwen2.5-VL-7B-Instruct \cite{wang2024qwen2}& 41.00 & 42.86 & 37.84 & 42.57 & 46.23 & 42.06 & 35.44 \\
Llama-3.2-11B-Vision-Instruct & 30.50 & 26.47 & 30.50 & 20.30 & 42.92 & 30.56 & 32.07 \\

\hline

\multicolumn{8}{l}{\textbf{Spatial Reasoning Models}} \\

SpaceOm \cite{chen2024spatialvlm}& 41.36 & 42.44 & 38.61 & 48.02 & 37.74 & 42.86 & 39.24 \\
SpaceThinker-Qwen2.5VL-3B \cite{chen2024spatialvlm}& 40.64 & 40.34 & 37.84 & 47.03 & 38.21 & 43.25 & 37.97 \\
SpaceQwen2.5-VL-3B-Instruct \cite{chen2024spatialvlm}& 40.14 & 31.51 & 35.14 & 37.62 & 37.74 & 50.79 & \textbf{47.26} \\
RoboPoint-vicuna-v1.5-7B-lora \cite{cai2025spatialbot}&38.00& 38.65 & 34.74 & 30.69 & 47.64 & 36.11 & 40.50  \\
RoboPoint-vicuna-v1.5-13B \cite{cai2025spatialbot}& 38.64 & 42.01 & 33.97 & 32.67 & 44.33 & 36.90 & 42.19 \\
SpatialLadder-3B \cite{li2025spatialladder}& 35.28 & 39.91 & 34.61 & 39.90 & 38.20 & 25.39 & 33.19\\
\hline

\multicolumn{8}{l}{\textbf{Backbone and Our Variants}} \\

\algo-4B & 44.71 & \textbf{46.22} & 46.33 & 42.57 & 53.77 & 36.51 & 43.88 \\
Qwen3-VL-8B-Thinking \cite{bai2025qwen3} (Base) & 44.78 & 42.01 & 47.10 & 47.02 & 51.88 & 40.47 & 41.35 \\
Qwen3-VL-8B-Thinking + GRPO& 46.21 & 43.70& 47.88 & 47.52 & 54.25& 41.27& 43.88\\
\algo-8B & \textbf{47.91} & 45.38 & \textbf{48.65} & \textbf{47.52} & \textbf{58.49} & 41.67 & 47.22 \\
\midrule
\bottomrule

\end{tabular}
}
\label{tab:table2}
\end{table*}

\subsection{Implementation Details}
\label{sec:implementation}
We instantiate \algo\ with Qwen3-VL-Thinking~\cite{bai2025qwen3}
backbones: \algo-8B uses Qwen3-VL-8B-Thinking, while the lighter
\algo-4B variant uses Qwen3-VL-4B-Thinking.
Both are post-trained with LoRA~\cite{ding2023} on all language-model linear
layers (\(r=32\), \(\alpha=64\), dropout \(0.05\)) and the vision encoder
(\(r=4\), \(\alpha=8\)). We train for 3 epochs on the
OmniSpatial~\cite{jia2025omnispatial} training split using two H100 GPUs,
an effective batch size of 32, and \(G=4\) rollout generations. We use AdamW
(\(5\times10^{-5}\), cosine schedule, 5\% warmup, \(\eta=0.01\)).
Pseudo-GT BBoxes are precomputed offline with Grounding DINO~\cite{liu2024grounding}.
We manually audit these BBoxes and test robustness to corruption
(\Cref{app:pseudo_gt_construction}). Reward hyperparameters are
\(\lambda_{\mathrm{fmt}}=0.2\), \(\lambda_s=1.0\), \(\gamma=0.3\), \(\beta=0.1\),
\(\alpha=2.0\), \(w_{\mathrm{iou}}=0.8\), and \(w_{\mathrm{label}}=0.2\).
Full details appear in \Cref{sec:hyperparameters}.

\subsection{Baselines}
\label{sec:baselines}
We compare \algo\ against the Qwen3-VL-8B-Thinking backbone, its standard
GRPO-trained variant, and open-source and specialized spatial reasoning
LVLMs; proprietary models serve as references. We evaluate on the held-out
OmniSpatial~\cite{jia2025omnispatial} test set (4 reasoning dimensions,
50 subcategories) and zero-shot on SpatiaLab~\cite{wasi2026spatialab}
(6 categories, 30 task types in naturalistic scenes) to assess transfer
across task designs and visual contexts. Details appear in
\Cref{app:benchmark_details}.

\begin{figure*}[t]
\centering
\includegraphics[width=1.0\textwidth]{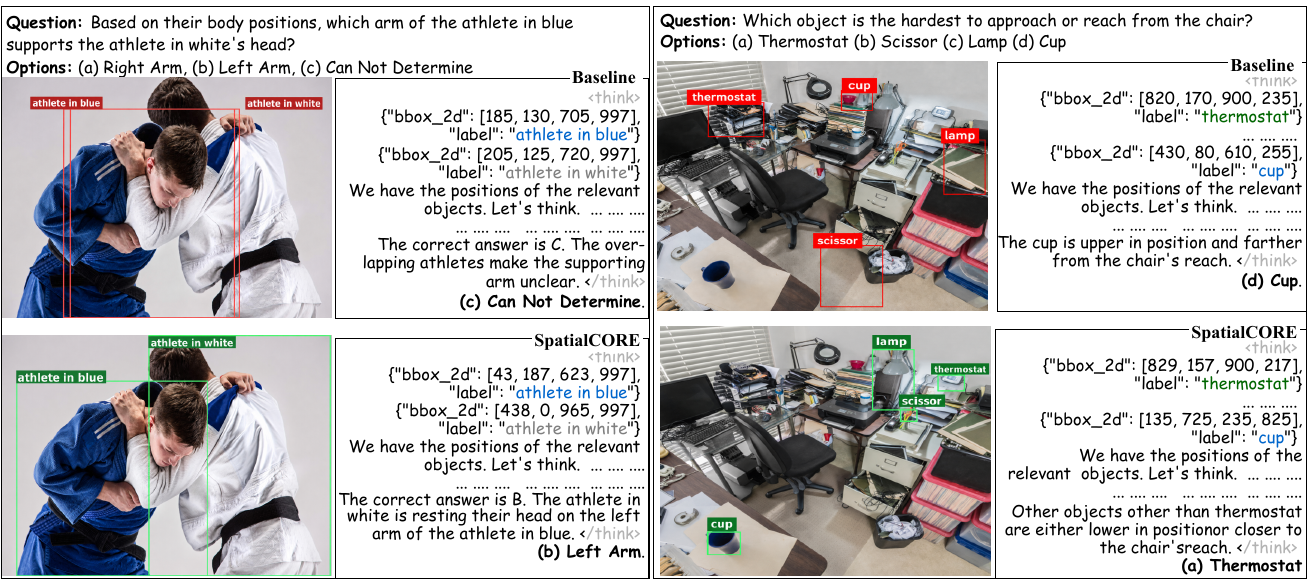}
\caption{\small Qualitative examples from OmniSpatial~\cite{jia2025omnispatial}
(left) and SpatiaLab~\cite{wasi2026spatialab} (right). In both cases, the
baseline (Qwen3-VL-8B-Thinking \cite{bai2025qwen3}) produces predicted BBoxes that mislocalize the task-relevant
objects, leading to incorrect or poorly grounded answers. \algo\ generates
more accurate predicted bounding boxes and reaches the correct final answer.
Bounding boxes are overlaid for visualization; full reasoning traces are in
the Appendix \Cref{app:qualitative}.}

\label{fig:figure3}
\vspace{-20pt}
\end{figure*}

\subsection{Results}

\noindent \textbf{Spatial Reasoning on Diverse and Challenging Tasks.}
\Cref{tab:table1} shows that \algo-8B achieves the highest weighted-average
accuracy among open-source and specialized spatial reasoning models on
OmniSpatial, surpassing the GRPO-trained backbone and showing particularly
strong gains in categories requiring object-centric spatial comparison.
\begin{wrapfigure}{t}{0.45\textwidth}
    \centering
    \vspace{-5mm}
    \includegraphics[width=\linewidth]{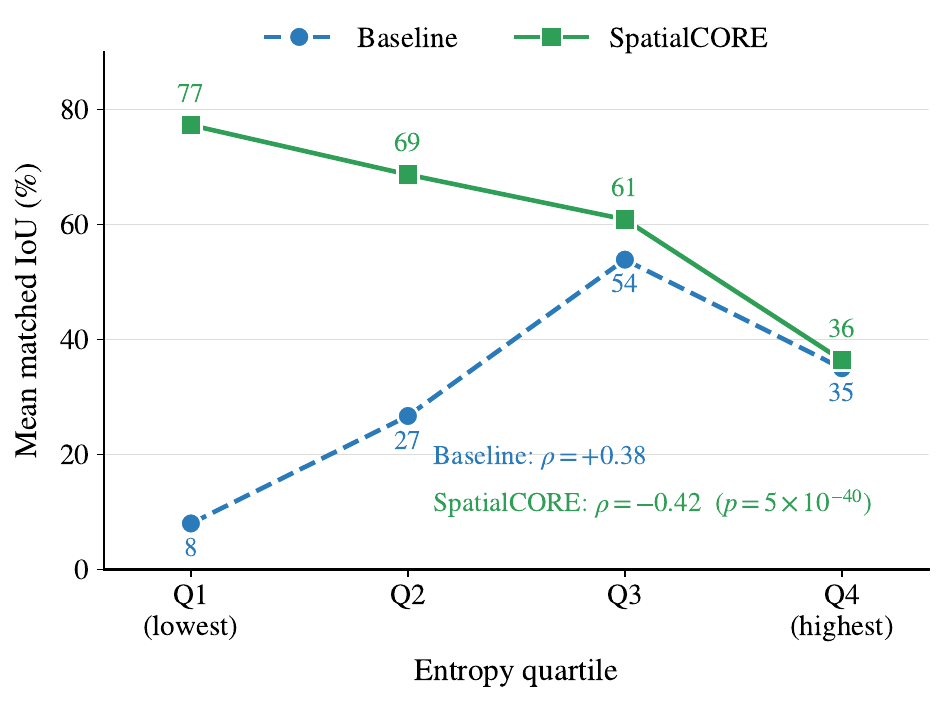}
    \vspace{-20pt}
    \caption{\small Mean matched BBox IoU across model-specific coordinate-token
entropy quartiles for the unadapted Qwen3-VL-8B-Thinking backbone (Baseline)
and \algo-8B. Lower entropy corresponds to more accurate grounding after
\algo\ post-training, whereas the baseline shows no consistent relationship.}

    \label{fig:figure4}
    \vspace{-10pt}
\end{wrapfigure}
Against specialized spatial
reasoning models, the margins are particularly large: \algo-8B outperforms
VST-RL-7B by $7.37\%$ and SpaceThinkerQwen2.5VL-3B by $8.04\%$, suggesting that
confidence-aware grounding provides a stronger training signal than
grounding supervision alone. Against open-source LVLMs, \algo-8B outperforms
SoFar-Qwen2.5VL-3B by $3.32\%$, InternVL3-14B by $2.52\%$, and Gemma-3-12B by
$4.75\%$.
The $4.56\%$ gain over its own backbone ($2.54\%$ gain over the trained backbone) demonstrates the effectiveness of \algo, while the matched GRPO ablation in Table~3 isolates the contribution of the spatial reward. Gains are strongest in \textit{traffic analysis} and \textit{localization}, tasks that require simultaneously comparing the positions of multiple task-relevant objects, where
confident BBoxes provide direct spatial evidence for the final
answer.
\textit{Allocentric} and \textit{hypothetical reasoning} show smaller gains, as they require
reasoning across multiple viewpoints, an input-level limitation that predicted
BBoxes from a single egocentric view cannot address. Notably, \algo-8B reaches the performance range of proprietary models as an
open-source system, and \algo-4B remains competitive at $44.68\%$ average accuracy.

\noindent \textbf{Zero-Shot Transfer to Unseen Distributions.}
On SpatiaLab (\Cref{tab:table2}), \algo-8B leads open-source and specialized
models on average and also outperforms the GRPO-trained backbone under
zero-shot evaluation. The improvements
over specialized models are substantial: $12.63\%$ over SpatialLadder-3B,
$7.27\%$ over SpaceThinker-Qwen2.5VL-3B, and $6.55\%$ over SpaceOm, with
\algo-8B also surpasses larger open-source LVLMs such as InternVL3.5-4B and
Qwen2.5-VL-7B-Instruct.
The $3.2\%$ gain over its own backbone ($1.7\%$ gain over the trained backbone) confirms that confidence-aware grounding transfers to benchmarks with different task designs and visual
distributions.
Gains are strongest in \textit{Relational Positioning}, where confident predicted BBoxes over multiple task-relevant objects provide direct evidence for spatial
comparisons. \textit{Size and Scale} estimation show smaller gains, suggesting that confidence-aware grounding primarily benefits object-centric spatial comparisons, while
scene-level scale estimation remains a distinct challenge. Notably, on average, \algo-8B matches
the performance range of proprietary systems on a fully unseen benchmark,
suggesting that the self-regulating spatial reward instills a grounding behavior
that generalizes beyond the training distribution.

\begin{wrapfigure}{t}{0.45\textwidth}
    \centering
    \includegraphics[width=\linewidth]{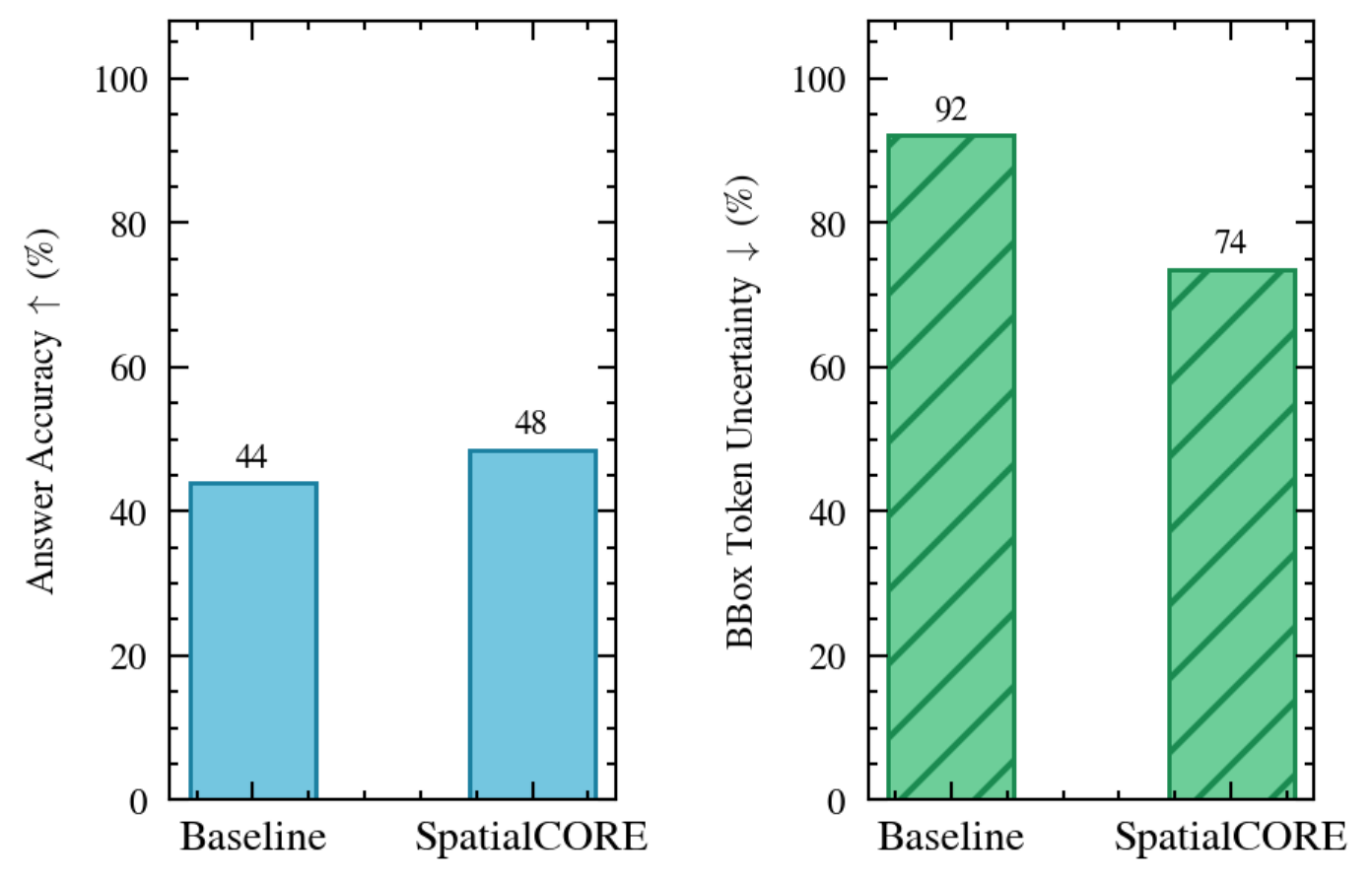}
    \vspace{-20pt}
    \caption{\small Bar plots comparing answer accuracy and predicted BBox
coordinate-token uncertainty for the unadapted Qwen3-VL-8B-Thinking
backbone (Baseline) and \algo-8B. Lower uncertainty indicates greater
confidence in BBoxes generated during reasoning.}
    \label{fig:figure5}
    \vspace{-10pt}
\end{wrapfigure}
\noindent \textbf{Qualitative Analysis.}
\Cref{fig:figure3} shows representative examples from OmniSpatial and SpatiaLab.
In both cases, the baseline produces grounding that fails to support the correct spatial decision, while \algo\ generates more accurate BBoxes and leverages them to reach the correct final answer. Notably, the examples reveal distinct failure modes: overlapping BBoxes can obscure relative-position reasoning, while mislocalized objects can corrupt reachability judgments.

\noindent \textbf{Learning to Ground with Confidence.}
The \textit{self-regulating spatial reward} reinforces grounding according
to both its localization quality and the model's confidence, encouraging
confident predictions where task-relevant objects are accurately localized.
This distinction matters because the baseline often assigns low uncertainty
to poorly localized BBoxes (\Cref{fig:figure4}). After \algo\
post-training, the most confident grounding achieves the highest
localization quality: mean matched IoU against pseudo-GT BBoxes reaches
$77\%$ in the lowest-uncertainty quartile and decreases consistently to
$36\%$ in the highest. The entropy--IoU correlation correspondingly shifts
from $\rho=+0.38$ to $\rho=-0.42$, making lower uncertainty a stronger
indicator of accurate localization. This change accompanies improved
spatial reasoning on the analyzed OmniSpatial test samples: answer
accuracy rises from $44\%$ to $48\%$, while BBox coordinate uncertainty
falls from $92\%$ to $74\%$ relative to
Qwen3-VL-8B-Thinking~\cite{bai2025qwen3} (\Cref{fig:figure5}).

\begin{wraptable}{r}{0.55\textwidth}
\vspace{-4mm}
\centering
\small
\setlength{\tabcolsep}{4pt}
\caption{\small Ablation study on the spatial interaction subset of
OmniSpatial~\cite{jia2025omnispatial}. Average is weighted by category
sample size. Best result is \textbf{bold}.}
\resizebox{\linewidth}{!}{
\begin{tabular}{lcccc}
\toprule
Variant & Avg. & \shortstack[c]{Traffic\\Anal.} &
\shortstack[c]{Loca-\\lization} & \shortstack[c]{Geospa.\\Strategy} \\
\midrule
\algo-8B\ (Full) & \textbf{56.33} & \textbf{54.11} & \textbf{62.85} & \textbf{51.81} \\
\quad w/o vision LoRA & 53.00 & 50.59 & 58.10 & 50.00 \\
\quad w/o conf. weighting & 52.33 & 49.41 & 58.09 & 49.09 \\
\quad w/o answer gate & 54.67 & 52.94 & 60.00 & 50.91 \\
\quad w/o pseudo-GT validity & 53.33 & 51.76 & 59.05 & 49.09 \\
\bottomrule
\end{tabular}}
\label{tab:ablation}
\vspace{-4mm}
\end{wraptable}
\noindent \textbf{Ablation Study.}
\Cref{tab:ablation} tests the defining feature of the
\textit{self-regulating spatial reward}: weighting BBox localization quality
by confidence in the generated grounding. In separate matched GRPO runs on
OmniSpatial's Spatial Interaction subset, \algo\ reaches 56.33\% accuracy;
removing confidence weighting while retaining the localization reward lowers
it to 52.33\%, the largest individual drop. This four-point gap shows the
value of learning from confidence beyond rewarding localization alone.
Vision LoRA and pseudo-GT validity contribute 3.33 and 3.00 points,
respectively, while the answer gate contributes 1.66 points by tying
rewarded grounding to final-answer correctness.

\section{Conclusion}
Spatial reasoning in LVLMs has largely focused on correct answers or accurate
localization, overlooking confidence in generated grounding. \algo\ uses this
confidence as a learning signal through a self-regulating spatial reward.
Benchmark gains and controlled ablations demonstrate its value; grounding
analyses show stronger alignment between confidence and localization quality.
Together, these findings advance confidence-aware spatial reasoning: learning
not only to produce grounding, but to reason from it with confidence.

\noindent \textbf{Limitations and Future Work}
\algo\ modifies the BBox-based post-training objective, leaving the backbone
and input representation unchanged. Future work could add depth, multi-view
context, or geometry-enhanced encoders to address 3D and non-boxable spatial
concepts.

\bibliography{iclr2027_conference}
\bibliographystyle{iclr2027_conference}
\newpage
\appendix
\section{Appendix: Training Configuration and Algorithm}
\label{sec:rationale}

\subsection{Hyperparameter Settings}
\label{sec:hyperparameters}

\Cref{tab:hyperparams} reports the main hyperparameters used for \algo-8B post-training, including the GRPO training setup, optimization settings, and reward configuration.

\begin{table}[t]
\centering
\caption{\small Hyperparameter settings for post-training \algo-8B with GRPO, including model configuration, training setup, optimization parameters, and reward weights.}
\label{tab:hyperparams}
\small
\setlength{\tabcolsep}{5pt}
\renewcommand{\arraystretch}{1.05}
\begin{tabularx}{\linewidth}{@{}l l X@{}}
\toprule
\textbf{Hyperparameter} & \textbf{Value} & \textbf{Notes} \\
\midrule

\multicolumn{3}{@{}l}{\textbf{\textit{Model and architecture}}} \\
Base model            & Qwen3-VL-8B-Thinking & Reasoning backbone \\
LoRA rank             & 32                   & Language-side LoRA \\
LoRA alpha            & 64                   & Language-side LoRA \\
LoRA dropout          & 0.05                 & -- \\
LoRA target modules   & Default              & Architecture defaults \\
Precision             & BF16                 & -- \\
Attention             & Flash Attention 2    & -- \\

\midrule
\multicolumn{3}{@{}l}{\textbf{\textit{Training}}} \\
Number of GPUs              & 2              & H100 GPUs \\
Epochs                      & 3              & -- \\
Per-device batch size       & 2              & -- \\
Rollout generations         & 4              & GRPO group size $G$ \\
Gradient accumulation steps & 8              & -- \\
Effective batch size        & 32             & $2 \times 2 \times 8$ \\
Total rollouts per update   & 128            & $32 \times 4$ generations \\
Max completion length       & 3{,}072 tokens & -- \\

\midrule
\multicolumn{3}{@{}l}{\textbf{\textit{Optimization}}} \\
Optimizer        & AdamW            & -- \\
Learning rate    & $5\times10^{-5}$ & -- \\
LR scheduler     & Cosine           & Minimum LR $=5\times10^{-6}$ \\
Warmup ratio     & 0.05             & -- \\
KL coefficient   & 0.01             & KL to reference policy \\

\midrule
\multicolumn{3}{@{}l}{\textbf{\textit{Reward}}} \\
Answer reward weight       & 1.0     & -- \\
Format reward weight       & 0.2     & -- \\
Grounding reward weight    & 1.0     & -- \\
Spatial answer gate        & 0.3     & Spatial reward scale for incorrect answers \\
Uncertainty floor          & 0.1     & Floor in confidence weighting $\omega$ \\
Grounding F-beta           & 2.0     & Coverage-weighted harmonic mean \\
IoU weight                 & 0.8     & Pairwise matching score \\
Label weight               & 0.2     & Pairwise matching score \\
Over-prediction penalty    & 0.3     & Per unmatched predicted BBox \\
BBox attempt bonus         & 0.05    & Added to format reward per valid grounding attempt \\
Uncertainty weighting      & Enabled & Active from step 0 \\

\bottomrule
\end{tabularx}
\end{table}

\subsection{Training Algorithm}
Algorithm~\ref{alg:spatialcore} summarizes one training iteration of \algo.

\begin{algorithm}[H]
\caption{Training \algo\ with self-regulating spatial reward}
\label{alg:spatialcore}
\begin{algorithmic}[1]
\Require LVLM policy $\pi_\theta$, reference policy $\pi_{\mathrm{ref}}$, training set $\mathcal{D}$, rollout size $G$, precomputed pseudo-GT grounding BBoxes
\For{each training step}
    \State Sample batch $\mathcal{B}\subset\mathcal{D}$
    \For{each $(I,q,y)\in\mathcal{B}$}
        \State Retrieve pseudo-GT BBoxes $\{(b_k^{\mathrm{gt}},\ell_k^{\mathrm{gt}},v_k)\}$
        \State Rollout: $\{o_i\}_{i=1}^{G} \sim \pi_\theta(\cdot \mid I,q)$
        \For{each trajectory $o_i$}
            \State Generated grounding: parse $\{b_j,\ell_j,\mathcal{S}_{i,j}\}$ from $o_i$
            \State Matching: compute pairwise $R_{\mathrm{BBox}}^{(j,k)}$ and $\mathcal{M}_i^*$
            \State Uncertainty: compute $H_{i,j}$ from BBox coordinate tokens
            \State Confidence: compute $C_{i,j}=1-H_{i,j}$ and $\omega_{i,j}$
            \State Spatial reward: compute $R_{\mathrm{spatial}}^{(i)}$ from $R_{\mathrm{BBox}}^{(j,k)}$, $\mathcal{M}_i^*$, and $\omega_{i,j}$
            \State Format/answer rewards: compute $R_{\mathrm{fmt}}^{(i)}$ and $R_{\mathrm{ans}}^{(i)}$
            \State Adaptive reward: compose $r_i$ from $R_{\mathrm{ans}}^{(i)}$, $R_{\mathrm{fmt}}^{(i)}$, and answer-gated $R_{\mathrm{spatial}}^{(i)}$
        \EndFor
        \State Advantage: compute $\{A_i\}_{i=1}^{G}$ from $\{r_i\}_{i=1}^{G}$
    \EndFor
    \State Update: optimize $\theta$ with GRPO and KL regularization to $\pi_{\mathrm{ref}}$
\EndFor
\end{algorithmic}
\end{algorithm}

\subsection{Pseudo-GT Supervision and Reliability}

\paragraph{BBox Construction.}
\label{app:pseudo_gt_construction}

The spatial reward compares generated BBoxes with reference locations for
task-relevant objects. We construct these pseudo-GT annotations offline,
before policy optimization. Each retained annotation is represented as
$(b_k^{\mathrm{gt}}, \ell_k^{\mathrm{gt}}, v_k)$, where
$b_k^{\mathrm{gt}}$ is a BBox in the LVLM's normalized image coordinates,
$\ell_k^{\mathrm{gt}}$ is its object label, and $v_k\in[0,1]$ is the
pseudo-GT validity used in reward computation.

We first identify the entities to localize. Given an image-question pair,
a fixed GPT-4o-mini~\cite{hurst2024gpt} prompt extracts physical objects
and persons mentioned in the question and answer options. It preserves
modifiers that distinguish instances, such as \texttt{boy in orange clothes},
\texttt{blue gear}, and \texttt{gray vehicle}. We filter non-visual terms
and relational or directional expressions, including \texttt{left},
\texttt{right}, \texttt{front}, \texttt{distance}, and \texttt{direction},
because they do not define object BBoxes.
\begin{wrapfigure}{t}{0.45\textwidth}
    \centering
    \includegraphics[width=\linewidth]{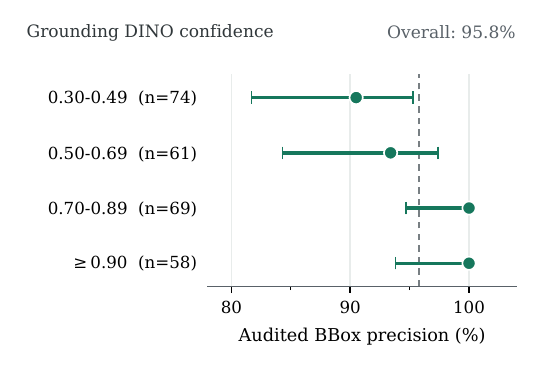}
    \vspace{-20pt}
    \caption{\small Manual audit precision of pseudo-GT BBoxes across Grounding
DINO confidence ranges. Points show precision and whiskers show 95\%
confidence intervals; labels give audited BBox counts. The dashed line
marks overall precision.}
    \label{fig:figure6}
    \vspace{-10pt}
\end{wrapfigure}
When no suitable object or person is identified, the phrase list remains
empty. Otherwise, the extracted phrases become queries for the grounding
model.

We then localize those phrases with Grounding DINO~\cite{liu2024grounding}.
For each sample with non-empty queries, we concatenate the phrases into a
text prompt and run zero-shot detection on the image. From the candidate
BBoxes, labels, and detection scores, we retain the highest-scoring BBox
per returned label and discard detections below the $0.3$ confidence
threshold. BBox coordinates are scaled to the LVLM's $[0,1000]$
image-coordinate convention, and each retained detection score becomes
its validity $v_k$. We omit pseudo-GT construction for task types where
object-level generated grounding is not meaningful. Finally, we save the
sample-level phrases and BBoxes in a JSON file for reward computation.

\paragraph{Quality Audit.}
\label{app:pseudo_gt_quality}
The spatial reward uses pseudo-GT BBoxes as localization targets, so their
reliability matters to the learning signal. We manually audited 262 retained
BBoxes from 200 randomly sampled training examples, counting a BBox as correct
if it localized the intended task-relevant object. Overall, 251 were correct
($95.8\%$). All 11 observed errors occurred in the two lower Grounding DINO
confidence ranges; every audited BBox with confidence at least $0.70$ was
correct (\Cref{fig:figure6}). This pattern motivates retaining detector
confidence as pseudo-GT validity $v_k$, giving less reliable references less
influence on the spatial reward. The matched ablation reinforces this choice:
removing validity weighting reduces accuracy from $56.33\%$ to $53.33\%$
(\Cref{tab:ablation}).

\paragraph{Pipeline Coverage.}
\label{app:pseudo_gt_coverage}

We trace pseudo-GT construction across the full OmniSpatial training split.
Of 5,643 samples containing BBox-localizable task-relevant objects, phrase
extraction yields queries for 4,063, and 4,007 receive at least one valid
Grounding DINO BBox above the $0.3$ confidence threshold. The resulting
coverage is $71.0\%$ of these samples and $98.6\%$ of those with phrase
queries. Samples without a valid reference often belong to Complex Logic or
Dynamic Reasoning tasks, where paths, sequences, or spatial relationships
may matter more to the answer than object localization alone.

The spatial reward applies only when grounding can be evaluated against a
valid reference. If no pseudo-GT BBox is available, the spatial reward is
zero; training continues with the answer and format rewards, and generated
BBoxes are neither rewarded nor penalized. If a valid reference is available
but the model generates no BBox, it receives a negative spatial reward with
coefficient $0.3$, scaled by pseudo-GT validity. This distinguishes the
absence of a reference from failure to ground an available target.

\paragraph{Robustness to Corrupted Pseudo-GT.}
\label{app:pseudo_gt_robustness}

The quality audit assesses the pseudo-GT BBoxes produced by our pipeline. To
test how strongly \algo\ depends on their accuracy, we retrain \algo-8B after
corrupting $40\%$ of valid pseudo-GT BBoxes. Each selected BBox is randomly
removed, assigned the label of a different valid object in the same sample, or
displaced until its IoU with the original BBox falls below $0.6$. These
interventions either remove spatial supervision or introduce a misleading
grounding target. We evaluate the resulting model on the full OmniSpatial
test set.

\begin{wraptable}{r}{0.56\columnwidth}
\centering
\caption{\small Robustness to pseudo-GT corruption on the full OmniSpatial
test set. Base denotes Qwen3-VL-8B-Thinking.}
\label{tab:table5}
\footnotesize
\setlength{\tabcolsep}{3pt}
\begin{tabular}{@{}lr@{}}
\toprule
Training setting & Acc. (\%) \\
\midrule
Base & 43.90 \\
\algo-8B ($40\%$ corrupted) & 46.70 \\
\algo-8B (original) & \textbf{48.46} \\
\bottomrule
\end{tabular}
\end{wraptable}

Even with two in five reference BBoxes corrupted, \algo-8B reaches $46.70\%$
accuracy, remaining $2.80$ percentage points above the unadapted backbone
(\Cref{tab:table5}). Relative to training with original pseudo-GT, accuracy
falls by $1.76$ points, showing that reference quality contributes to the
gain. At the same time, the improvement persists when spatial supervision is
partly missing or misleading: $60\%$ of the BBoxes remain intact, and the
answer and format rewards continue to provide learning signals for every
sample. Together with the quality audit, this experiment establishes both
the value of reliable pseudo-GT and the resilience of \algo\ to substantial
corruption of its grounding targets.

\subsection{Reward and Implementation Details}

\paragraph{Pseudo-GT Matching.}
\label{app:pseudo_gt_matching}

Generated grounding in a rollout may not align one-to-one with the precomputed
pseudo-GT BBoxes. The policy can produce a different number of BBoxes than the
pseudo-GT set, and its labels may use different but compatible wording. For
example, a generated label \texttt{black sedan ahead} should still match a
pseudo-GT label \texttt{black sedan}. Conversely, geometric overlap alone is
insufficient: a predicted BBox labeled \texttt{traffic sign} may overlap a
pseudo-GT BBox for \texttt{traffic light}, but the semantic mismatch should make
this assignment weaker. We therefore match predicted and pseudo-GT BBoxes using
both geometric overlap and label similarity.

For a predicted BBox with object label $(b_j,\ell_j)$ and a pseudo-GT BBox
$(b_k^{\mathrm{gt}},\ell_k^{\mathrm{gt}},v_k)$, we define the pairwise BBox
reward as
\begin{equation}
R_{\mathrm{BBox}}^{(j,k)} =
\left(
w_{\mathrm{iou}}
\max\left(0,\mathrm{IoU}(b_j,b_k^{\mathrm{gt}})-\tau_{\mathrm{iou}}\right)
+
w_{\mathrm{label}}\,\mathrm{Sim}(\ell_j,\ell_k^{\mathrm{gt}})
\right)v_k ,
\end{equation}
where $w_{\mathrm{iou}}+w_{\mathrm{label}}=1$ and $\tau_{\mathrm{iou}}$ is an
IoU margin. The clipped IoU term suppresses weak geometric overlap, while the
label-similarity term favors assignments between semantically compatible object
phrases. This prevents overlapping but mismatched BBoxes from being treated as
strong matches.

We compute label similarity through semantic label representations i.e., a lightweight \textit{bag-of-words}
representation. Each label is tokenized into words and represented by a binary
word-presence vector. For example, \texttt{black sedan ahead} and
\texttt{black sedan} share the key words \texttt{black} and \texttt{sedan},
yielding high cosine similarity, whereas \texttt{traffic sign} and
\texttt{traffic light} share only \texttt{traffic} and receive lower similarity.
Formally, let $\phi(\ell)$ denote the bag-of-words vector for label $\ell$. We
compute
\begin{equation}
\mathrm{Sim}(\ell_j,\ell_k^{\mathrm{gt}})
=
\frac{
\phi(\ell_j)^\top \phi(\ell_k^{\mathrm{gt}})
}{
\|\phi(\ell_j)\|_2\,\|\phi(\ell_k^{\mathrm{gt}})\|_2
}.
\end{equation}
This lexical similarity is sufficient for our setting because labels are short
object phrases extracted from questions, answer options, and generated
grounding. It also avoids adding an external embedding model to reward
computation.

The pseudo-GT validity $v_k$ scales the pairwise reward by the reliability of
the grounding-model localization. High-validity pseudo-GT BBoxes therefore have
a stronger effect on matching, while lower-validity or noisier BBoxes contribute
less to the spatial reward.

After computing all pairwise rewards, we use Hungarian
matching~\cite{kuhn1955hungarian} to obtain a maximum-reward one-to-one
assignment:
\begin{equation}
\mathcal{M}_i^*
=
\arg\max_{\mathcal{M}_i}
\sum_{(j,k)\in\mathcal{M}_i} R_{\mathrm{BBox}}^{(j,k)} .
\end{equation}
The one-to-one constraint prevents a single predicted BBox from matching
multiple pseudo-GT BBoxes and prevents multiple predictions from claiming the
same pseudo-GT BBox. The matched pairs in $\mathcal{M}_i^*$ are then used to
compute the confidence-weighted spatial reward.

\paragraph{Additional Reward Shaping.}
We use two lightweight shaping terms for training stability. First, the format
reward includes a BBox attempt bonus of $0.05$ when pseudo-GT BBoxes are
available and the trajectory contains at least one valid predicted BBox whose
label overlaps with the question or answer options. This reduces the incentive
to avoid BBox generation while filtering out irrelevant grounding attempts.
Second, the spatial reward applies an over-prediction penalty of $0.3$ to valid
predicted BBoxes that remain unmatched after one-to-one assignment with
pseudo-GT BBoxes, limiting extra generated grounding beyond the task-relevant
objects. These auxiliary terms stabilize the grounded format during
post-training, while the main spatial supervision is provided by the
confidence-weighted matching reward.

\paragraph{SFT Cold Start.}
\label{app:sft_cold_start}

Before GRPO post-training, we perform a supervised cold-start stage on
1,000 OmniSpatial~\cite{jia2025omnispatial} training samples. This stage
initializes the policy with the required grounded reasoning format and teaches
the distinction between boxable task-relevant objects and non-boxable spatial
terms. When reliable pseudo-GT BBoxes are available, the target completions
include BBox lines, followed by a reasoning trace and final answer in the same
format used during reinforcement learning. We train a LoRA adapter with
completion-only cross-entropy, masking prompt tokens while freezing the visual
encoder and updating only language-side adaptation parameters. The resulting
adapter initializes the GRPO policy, after which \algo\ optimizes the
self-regulating spatial reward described in the main method.

To assess the necessity of this initialization, we additionally train the base
model with GRPO for one full epoch without the 1,000-sample SFT cold start.
Without SFT, the model does not converge to the grounded format: the grounding
reward remains between $-0.125$ and $+0.003$, ends at $0.000$, and the model
ultimately stops generating BBoxes. Its general reasoning and final-answer
format remain stable, indicating that the collapse is specific to generated
grounding. In contrast, with the cold start, \algo\ maintains a positive
grounding reward between $0.15$ and $0.33$ and generates BBoxes in
$95$--$100\%$ of training responses throughout GRPO. These dynamics indicate
that the cold start primarily establishes the grounded output structure; once
this structure is maintained, the spatial reward drives subsequent improvement.

\paragraph{System Prompt Design.}
\label{app:system_prompt}

We use the fixed system prompt in \Cref{fig:figure7a} during training to enforce
a consistent grounded reasoning format. Since Qwen3-VL-Thinking
\cite{bai2025qwen3} automatically prefills the opening \texttt{<think>} token
after the user message, the prompt does not ask the model to generate
\texttt{<think>}; it only requires the model to close the reasoning segment with
\texttt{</think>}. The prompt further instructs the model to output relevant
BBoxes before the reasoning trace and produce a single final answer letter after
\texttt{</think>}, allowing the training pipeline to parse generated grounding,
reasoning, and final answers consistently across sampled trajectories.

\paragraph{Compute Resources.}
\label{app:compute}

The main training configuration is described in \Cref{sec:implementation} and
\Cref{tab:hyperparams}. The final \algo\ GRPO run was performed on 2 NVIDIA H100
GPUs using BF16 precision and vLLM-based rollout generation. Training required
approximately 31 hours, corresponding to about 62 H100 GPU-hours. This includes
rollout sampling, reward computation, and policy optimization for the final
reported model. Runtime may vary depending on the generation backend, rollout
length, decoding settings, batching efficiency, and system load. The reported
estimate is therefore intended as a practical reference for reproducing the main
training run rather than an exact hardware-independent cost. It excludes
preliminary debugging, hyperparameter exploration, failed runs, and additional
baseline or ablation experiments.

\subsection{Benchmark Details}
\label{app:benchmark_details}

\paragraph{OmniSpatial.}
OmniSpatial~\cite{jia2025omnispatial} contains more than 8.4K question--answer pairs covering four spatial
reasoning dimensions: dynamic reasoning, spatial interaction, complex spatial
logic, and perspective taking. These dimensions are further divided into 50
fine-grained task subcategories. We use the official training split of 6,902
samples for \algo\ post-training and evaluate on the official held-out test split
of 1,533 samples.

\paragraph{SpatiaLab.}
SpatiaLab~\cite{wasi2026spatialab} contains 1,400 visual question--answer pairs from realistic,
unconstrained scenes. It covers six spatial reasoning categories: relative
positioning, depth and occlusion, orientation, size and scale, spatial
navigation, and 3D geometry, with five task types per category. We evaluate
\algo\ in the multiple-choice setting. No SpatiaLab samples are used during
post-training.

We evaluate \algo\ on OmniSpatial~\cite{jia2025omnispatial} and
SpatiaLab~\cite{wasi2026spatialab}. OmniSpatial is used for post-training and
held-out evaluation, while SpatiaLab is used only for zero-shot evaluation.

\paragraph{Evaluation Protocol.}
For both benchmarks, we report accuracy. A prediction is counted as correct if
the selected option matches the ground-truth answer. Category-level scores are
computed over the samples in each category. Overall accuracy is computed over the
full evaluation set and is equivalently reported as a sample-weighted average
across categories.

\section{Appendix: Rationale for Confidence-Guided Grounding}
\label{app:theory}

We provide an optimization-based rationale for confidence-guided generated
grounding in \algo, using the GRPO policy optimization framework
\cite{shao2024deepseekmath}. Under answer-only rewards, trajectories that
produce the same final answer receive the same reward signal, even if their
generated grounding differs in confidence. This can make the objective
insensitive to whether predicted BBoxes are confidently localized. The
confidence-weighted spatial reward reduces this indifference by incorporating
predicted BBox coordinate-token uncertainty into the reward signal.

\subsection{Gradient Indifference Under Answer-Only Supervision}

Under standard GRPO, the policy objective over a group of $G$ trajectories is:
\begin{equation}
\label{eq:grpo_obj}
J(\theta) = \mathbb{E}_{(I,q) \sim \mathcal{D},\, \{o_i\} \sim \pi_{\theta_{\text{old}}}}
\left[
\frac{1}{G} \sum_{i=1}^{G}
\min\!\left(\rho_i(\theta) A_i,\;
\text{clip}(\rho_i(\theta), 1-\delta, 1+\delta) A_i\right)
- \eta\, D_{\mathrm{KL}}(\pi_\theta \| \pi_{\mathrm{ref}})
\right],
\end{equation}
where $\rho_i(\theta) = \pi_\theta(o_i \mid I, q) /
\pi_{\theta_{\text{old}}}(o_i \mid I, q)$ is the importance ratio and $A_i$ is
the group-relative advantage.

Under answer-only supervision, $r_i = R_{\mathrm{ans}}^{(i)}$, and the advantage
$A_i$ depends only on whether the final answer is correct. Consider two
trajectories $o_i$ and $o_{i'}$ that produce the same correct final answer but
differ in predicted BBox coordinate-token uncertainty: $o_i$ produces
low-uncertainty generated grounding $H_{i,j} \approx 0$, while $o_{i'}$ produces
high-uncertainty generated grounding $H_{i',j} \approx 1$. Since
$R_{\mathrm{ans}}^{(i)} = R_{\mathrm{ans}}^{(i')} = 1$, their answer-only
rewards are identical, and their group-relative advantages provide no preference
between confident and uncertain generated grounding. For the unclipped
policy-gradient term, coordinate tokens from these trajectories therefore receive
the same advantage whenever the final answers match:
\begin{equation}
\label{eq:grad_indiff}
A_i = A_{i'}
\quad \Rightarrow \quad
\nabla_\theta \log \pi_\theta(x_{i,t} \mid x_{i,<t}, I, q)\, A_i
\;\text{and}\;
\nabla_\theta \log \pi_\theta(x_{i',t} \mid x_{i',<t}, I, q)\, A_{i'}
\end{equation}
are weighted by the same trajectory-level advantage, for
$x_{i,t} \in \mathcal{S}_{i,j}$ and
$x_{i',t} \in \mathcal{S}_{i',j}$. Thus, answer-only supervision provides no
gradient signal that separates low-uncertainty generated grounding from
high-uncertainty generated grounding. As a result, uncertain generated grounding
may still be reinforced when it co-occurs with a correct final answer.

\subsection{Confidence-Weighted Rewards Break This Indifference}

\algo\ introduces a spatial reward $R_{\mathrm{spatial}}^{(i)}$ that depends on
the confidence weight
\begin{equation}
\omega_{i,j} = \beta + (1-\beta)(1 - H_{i,j}),
\end{equation}
where $H_{i,j}$ is the normalized predicted BBox coordinate-token uncertainty and
$\beta$ is the confidence floor. This makes the spatial reward sensitive to
generated grounding confidence. For trajectories with the same answer and format
rewards, the composite reward in \Cref{eq:reward_compose} can still differ
through the spatial term:
\begin{equation}
\label{eq:reward_diff}
r_i - r_{i'} =
\lambda_s \left(
g(R_{\mathrm{ans}}^{(i)}) R_{\mathrm{spatial}}^{(i)}
-
g(R_{\mathrm{ans}}^{(i')}) R_{\mathrm{spatial}}^{(i')}
\right).
\end{equation}
When two trajectories have comparable matched BBox quality but different
coordinate-token uncertainty, the lower-uncertainty trajectory receives a larger
confidence weight $\omega_{i,j}$ and therefore a larger spatial reward. This
creates an advantage gap that favors confident generated grounding over uncertain
generated grounding, even when both trajectories reach the same final answer.

\subsection{Answer Gate Preserves Generated Grounding Signal on Incorrect Trajectories}

A further concern is whether the spatial reward is lost when the final answer is
incorrect. Under answer-only supervision, incorrect trajectories receive
$r_i = 0$ regardless of whether their generated grounding is useful. \algo\
addresses this through the answer gate $g(\cdot)$ in \Cref{eq:reward_compose},
which keeps a reduced spatial reward when the final answer is incorrect:
\begin{equation}
\label{eq:wrong_reward}
r_i = \lambda_{\mathrm{fmt}} R_{\mathrm{fmt}}^{(i)}
+ \lambda_s \gamma R_{\mathrm{spatial}}^{(i)}, \quad
\text{if } R_{\mathrm{ans}}^{(i)} = 0,
\end{equation}
where $\gamma \in (0,1)$. Thus, trajectories with incorrect final answers can
still retain partial credit for useful generated grounding. This allows generated
grounding to improve before the model consistently predicts the correct final
answer.

\begin{figure}[p]
    \centering
    \includegraphics[width=\textwidth]{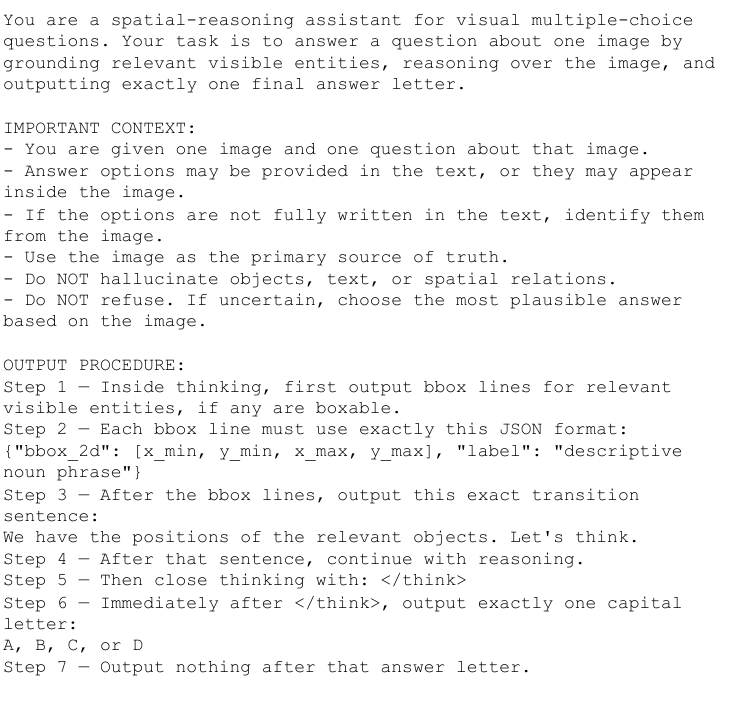}
    \caption{\small System prompt used to standardize generated grounding trajectories during \algo\ training. Continued on the next page.}
    \label{fig:figure7a}
\end{figure}

\begin{figure}[p]
    \ContinuedFloat
    \centering
    \includegraphics[width=\textwidth]{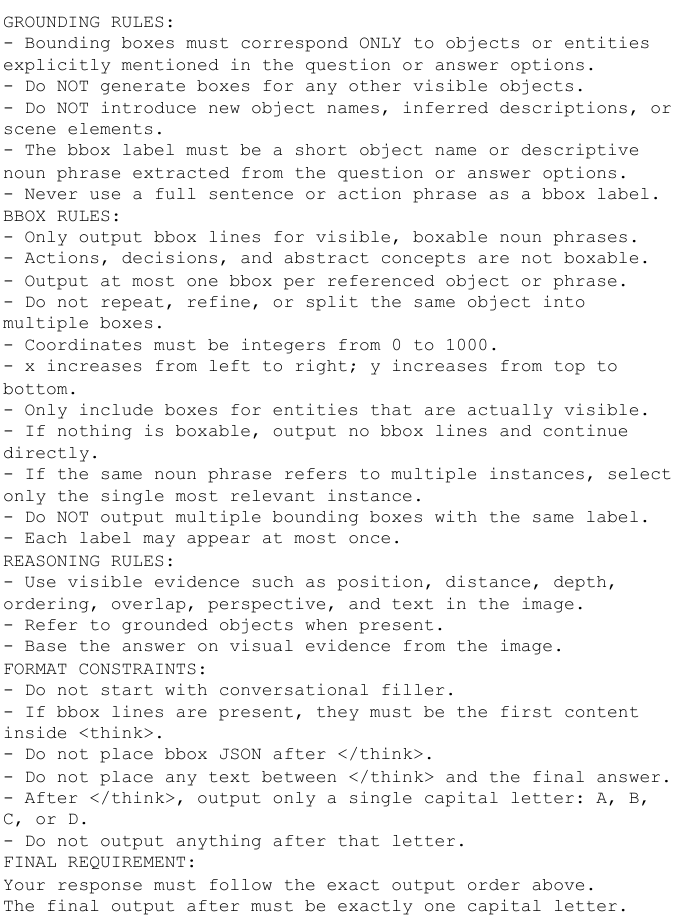}
    \caption{\small System prompt used to standardize generated grounding trajectories during \algo\ training, continued.}
\end{figure}

\section{Appendix: Additional Qualitative Results}
\label{app:qualitative}

\begin{figure*}[p]
\centering
\includegraphics[width=0.9\textwidth]{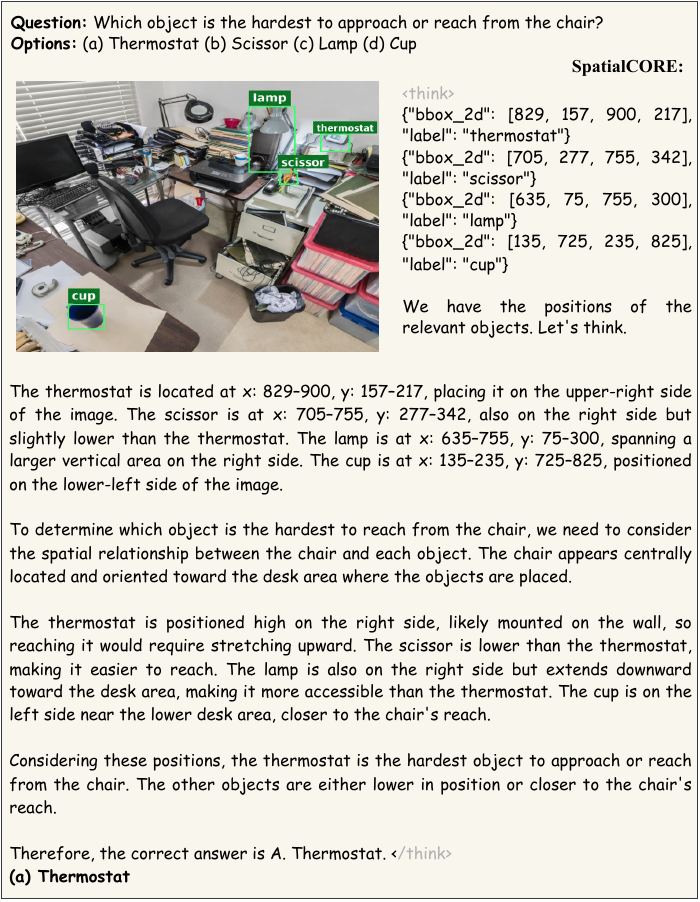}
\caption{\small Qualitative example of confidence-aware grounded spatial reasoning with \algo. The model first generates predicted BBoxes for the task-relevant objects, including \texttt{thermostat}, \texttt{scissor}, \texttt{lamp}, and \texttt{cup}, and then reasons over their localized positions relative to the chair. By producing confident generated grounding during the reasoning trace, \algo\ identifies the wall-mounted thermostat as the hardest object to reach and selects the correct final answer. Bounding boxes are overlaid only for visualization.}
\label{fig:figure8}
\vspace{-10pt}
\end{figure*}

\Cref{fig:figure8} presents a complete \algo\ trajectory. The example includes
the predicted BBoxes generated in the reasoning trace, the subsequent spatial
reasoning, and the final answer. It shows how \algo\ makes generated grounding
part of the reasoning process, using localized task-relevant objects as evidence
for the final spatial decision.

\FloatBarrier

\end{document}